\def\year{2018}\relax
\documentclass[letterpaper]{article} %DO NOT CHANGE THIS
\usepackage{aaai18}  %Required
\usepackage{times}  %Required
\usepackage{helvet}  %Required
\usepackage{courier}  %Required
\usepackage{url}  %Required
\usepackage{graphicx}  %Required

\usepackage{graphicx,float,url,multirow,makecell,booktabs,color,subfigure,caption,setspace}
\usepackage{amsmath}
\usepackage{times}
\usepackage{epsfig}
\usepackage{graphicx}
\usepackage{amsmath}
\usepackage{amssymb}

\begin{document}

\newcommand{\argmin}{\operatornamewithlimits{argmin}}
\newcommand{\argmax}{\operatornamewithlimits{argmax}}

\newcommand{\tabincell}[2]{\begin{tabular}{@{}#1@{}}#2\end{tabular}}

\newtheorem{theorem}{$\textbf{Theorem}$}
\newtheorem{lemma}[theorem]{$\textbf{Lemma}$}

% Strut macros for skipping spaces above and below text in tables.
\def\abovestrut#1{\rule[0in]{0in}{#1}\ignorespaces}
\def\belowstrut#1{\rule[-#1]{0in}{#1}\ignorespaces}
\def\abovespace{\abovestrut{0.01in}}
\def\belowspace{\belowstrut{-0.01in}}

\title{Generalization Tower Network: \\
A Novel Deep Neural Network Architecture for Multi-Task Learning}
%\author{AAAI Press\\
%Association for the Advancement of Artificial Intelligence\\
%2275 East Bayshore Road, Suite 160\\
%Palo Alto, California 94303\\
%}
\author{\textbf{Yuhang Song}$^1$, Mai Xu$^2$, Songyang Zhang$^3$, \textbf{Liangyu Huo}$^4$ \\
	$^{1}$\textbf{Beijing University of Aeronautics and Astronautics}\\
	yuhangsong@buaa.edu.cn, maixu@buaa.edu.cn, zhangsy1@shanghaitech.edu.cn, huoliangyu@buaa.edu.cn}
\maketitle
\begin{abstract}
Deep learning (DL) advances state-of-the-art reinforcement learning (RL), by incorporating deep neural networks in learning representations from the input to RL. However, the conventional deep neural network architecture is limited in learning representations for multi-task RL (MT-RL), as multiple tasks can refer to different kinds of representations. In this paper, we thus propose a novel deep neural network architecture, namely generalization tower network (GTN), which can achieve MT-RL within a single learned model. Specifically, the architecture of GTN is composed of both horizontal and vertical streams. In our GTN architecture, horizontal streams are used to learn representation shared in similar tasks. In contrast, the vertical streams are introduced to be more suitable for handling diverse tasks, which encodes hierarchical shared knowledge of these tasks. The effectiveness of the introduced vertical stream is validated by experimental results. Experimental results further verify that our GTN architecture is able to advance the state-of-the-art MT-RL, via being tested on 51 Atari games.
\end{abstract}

\section{Introduction}
Reinforcement learning (RL) is an active research area of artificial intelligence (AI). It aims at making a computer, as the agent, take actions in an environment to maximize the cumulative reward. The past decades have witnessed the evolution of RL \cite{sutton1998reinforcement}, especially the recent development of deep RL (DRL) \cite{mnih2015human}. The recent DRL almost achieves human-level intelligence in mastering single task, for a small yet growing set of scenarios. Unfortunately, DRL is still in its infancy for handling more than one task, i.e., multi-task RL (MT-RL) \cite{caruana1998multitask}.

For MT-RL, recent works have already achieved notable advances on inter-task transfer, showing the feasibility of transferring learned knowledge from one task to another. Actor-mimic \cite{parisotto16_actormimic} and policy distillation \cite{rusu2015policy} leverage techniques from model compression to perform state-of-the-art inter-task transfer. Beyond inter-task transfer, multi-task generalization has also been considered in MT-RL. For example, shared representations have been verified to be formable, when facing diverse tasks \cite{borsa2016learning,romoff2016deep}. The approach of progressive networks \cite{rusu2016progressive} is insightful to transfer knowledge from the hidden layers of previously learned model to learn a never-experienced task. Yet, its multi-task ability is accomplished by storing all learned models and relying on inference to specify a corresponding model when switching to a specific task.

Despite all recent achievements for MT-RL, the multi-task generalization is still limited, or requires all learned models to be stored and then be manually selected according to the task. To avoid such limitation, this paper proposes\footnote{Code to reproduce this work is publicly available online for
facilitating future research: https://github.com/YuhangSong/GTN.
} a novel deep neural network architecture for MT-RL, which is capable of handling all tasks within a single model. It is worth pointing out that our architecture can be easily combined with the existing MT-RL algorithms, e.g., actor-mimic \cite{parisotto16_actormimic}, policy distillation \cite{rusu2015policy}, progressive networks \cite{rusu2016progressive} and EWC \cite{kirkpatrick2017overcoming}, as our work mainly focuses on the architecture of deep neural network for MT-RL.
\begin{figure*}[htbp]
	\begin{center}
		\centerline{\includegraphics[width=0.9\textwidth]{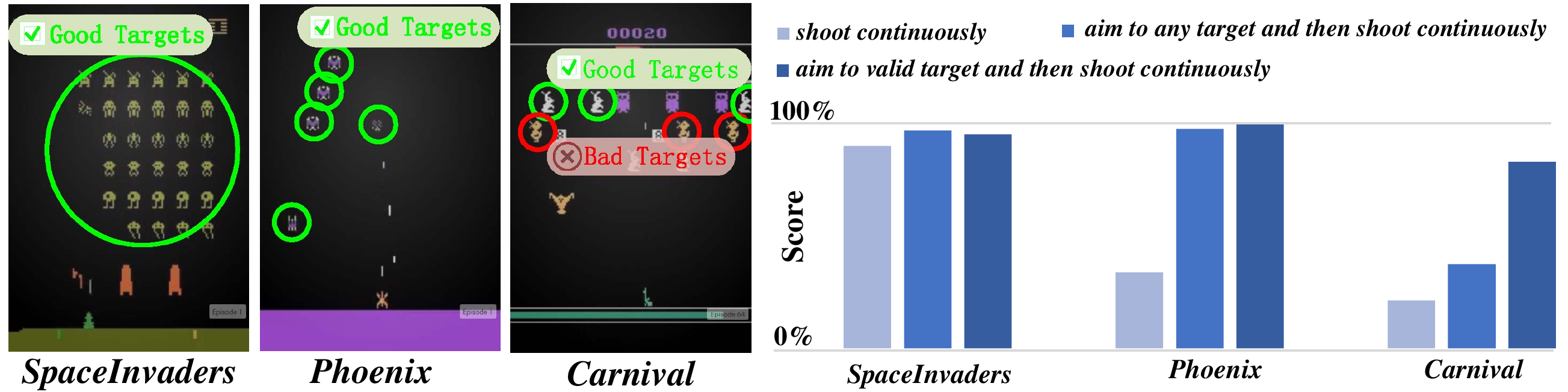}}%of
		\caption{\textit{Hierarchical shared knowledge} in shooting games. \textbf{\textit{SpaveInvaders}} can obtain high scores by knowledge \textit{shoot continuously}, since there are substantial and intensive targets. As for \textbf{\textit{Phoenix}}, the targets are much fewer but are all valid targets, high scores are obtained with knowledge \textit{aim to targets and then shoot continuously}. \textbf{\textit{Carnival}} is much more demanding for the players, since there are both bad targets and good targets. Thus, we should \textit{aim to valid targets and then shoot continuously}. Thus, all three shooting games share knowledge of \textit{shoot continuously}. Both \textbf{\textit{Phoenix}} and \textbf{\textit{Carnival}} games have knowledge of \textit{aim to targets and then shoot continuously}. Only \textbf{\textit{Carnival}} holds knowledge of \textit{aim to valid targets and then shoot continuously}.}
		\label{shoot}
	\end{center}
\end{figure*}
To be more specific, this paper proposes the generalization tower network (GTN), a novel deep neural network architecture for MT-RL.
The proposed GTN is inspired by our finding in Figure \ref{shoot}, which implies that similar tasks in MT-RL share some hierarchical common knowledge, called \textit{hierarchical shared knowledge}.
As shown in Figure \ref{shoot}, we find that all shooting games\footnote{For obtaining scores, we asked 10 volunteers aging from 18 to 26 to play these \textit{shooting games}. All of them are naive to these games, avoiding the influence of prior knowledge. The volunteers were told to play the games only with three kinds of knowledge: I. \textit{shoot continuously}, II. \textit{aim to targets and then shoot continuously} and III. \textit{aim to valid targets and then shoot continuously}. Scores of games are not displayed (by occlusion) for making knowledge valid.  With each knowledge, they play three shooting games in a random order, each for 90 seconds.} in Atari \cite{bellemare2013arcade} need to \textit{shoot continuously}, and some of them may further require \textit{shooting to targets} or even \textit{shooting to valid targets}.
Thus, knowledge needs to be encoded in hierarchy for different tasks of MT-RL.
Accordingly, our GTN architecture has both horizontal and vertical streams.
Horizontal streams learn to extract multi-layer representations from high-dimensional input, whereas vertical streams generalize \textit{hierarchical shared knowledge}.
The functionality of the introduced vertical steams is two-fold: (1) When facing multiple tasks, low-level features may be shared in the vertical stream; (2) Some tasks can be learned by short streams, while others may be learnt by long streams.
We further verify through experiments that vertical streams of GTN are capable of improving the performance of mastering multiple tasks within a single model for MT-RL.
More importantly, experimental results show that our GTN approach is better than the state-of-the-art asynchronous advantage actor-critic (A3C) for MT-RL \cite{romoff2016deep} on 51 games of Atari, with higher scores.

\section{Related Work}
\label{Related Work}
\begin{figure*}[htb]
	\begin{center}
		\centerline{\includegraphics[width=0.82\textwidth]{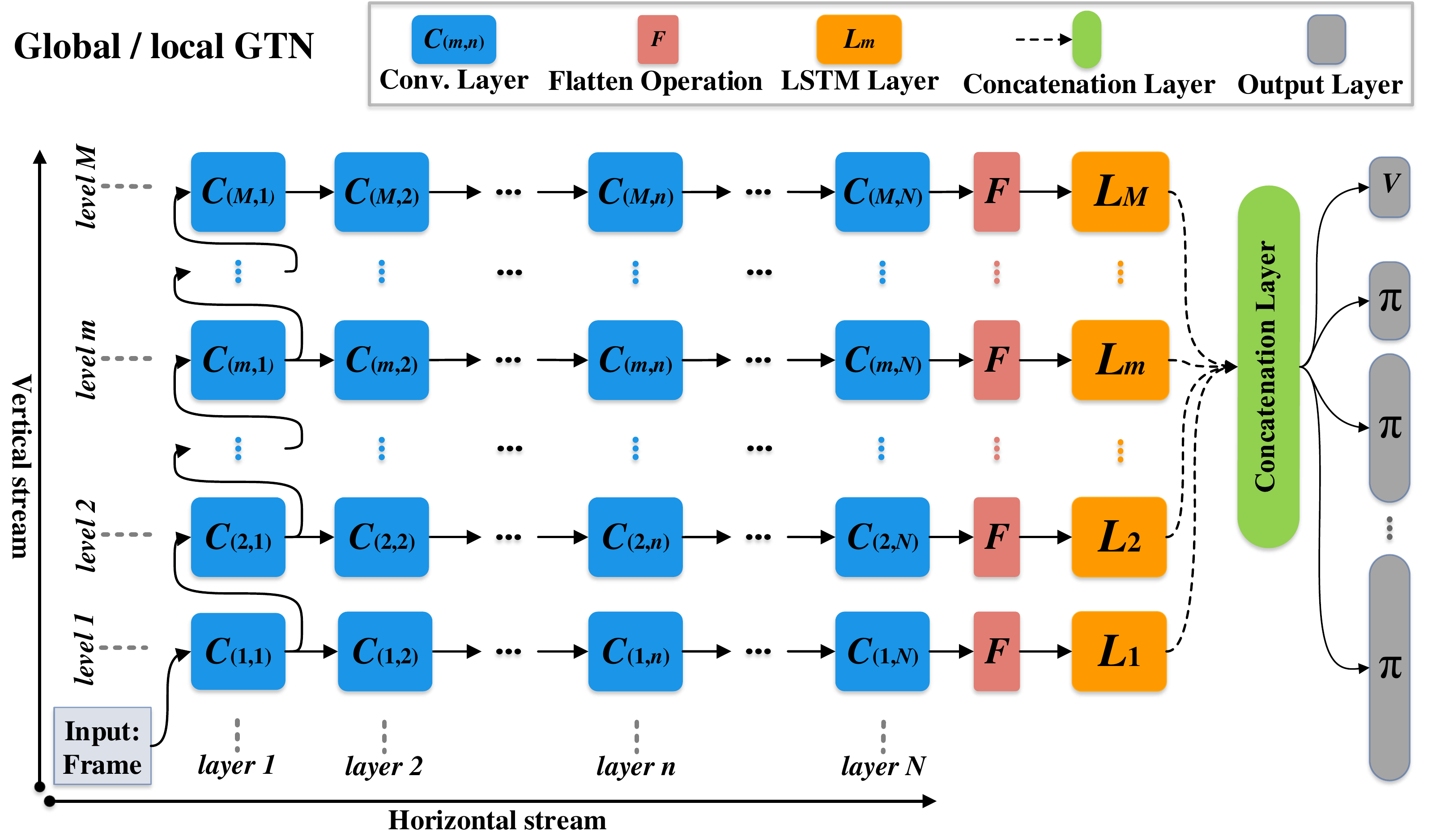}}
		\caption{
			Architecture of Generalization Tower Network (GTN).
			As shown in this figure, GTN contains both horizontal and vertical streams.
			Specifically, horizontal streams at different levels are similarly composed of convolutional layers, flatten layer and LSTM layer.
			As for the vertical streams, they are achieved by sharing layers of horizontal streams in hierarchy.
		}
		\label{figure-main-framework}
	\end{center}
\end{figure*}
\subsection{Deep Reinforcement Learning}
Taking advantage of recent success in deep learning (DL) \cite{lecun2015deep} and long-existing research in RL \cite{sutton1998reinforcement}, DRL has made great progress, starting from deep $Q$-learning \cite{mnih2015human}. Notable advances include double $Q$-learning \cite{van2016deep}, prioritized experience replay \cite{schaul2015prioritized}, dueling networks \cite{wang2016dueling}, and asynchronous methods \cite{mnih2016asynchronous}. Utilizing advances from these key achievements, A3C in \cite{mnih2016asynchronous} achieves the state-of-the-art performance in mastering RL tasks, reaching human level intelligence in various domains. However, in spite of the existing achievements, agents in above RL works are still limited to learning and mastering one task at a time.

\subsection{Multi-Task Reinforcement Learning}
MT-RL is crucial in the RL area, which learns more than one task at a time. In this regard, the ultimate goals for MT-RL can be classified as:
(1) \textbf{Inter-Task Transfer}, i.e., the RL model learned for one task can help in learning other tasks;
(2) \textbf{Multi-Task Generalization}, i.e.,  multiple tasks can be handled by learning a single RL model.
Above topics have long been reputed as a critical challenge in many AI works \cite{ring1994continual,silver2013lifelong,taylor2011introduction}. Recent advances in this topic can be generally divided into following two directions.

\paragraph{Inter-Task Transfer.}
Inter-Task transfer works primarily focus on how to transfer knowledge from the learned model to a new one, when facing a new task.
Upon such transfer, the new learned models are capable of mastering more tasks, in a way that different tasks are learned one by one.
However, the main challenge for inter-task transfer is \textit{catastrophic forgetting} \cite{mccloskey1989catastrophic,mcclelland1995there,ratcliff1990connectionist}, which means that the newly learned knowledge may completely break previous learned knowledge of old tasks. Some novel advances \cite{rusu2016progressive,fahlman1990cascade,kirkpatrick2017overcoming,rusu2015policy,terekhov2015knowledge,parisotto16_actormimic} are produced to this issue.

\paragraph{Multi-Task Generalization.} Multi-Task generalization \cite{borsa2016learning,romoff2016deep,sermanet2016unsupervised} generally concentrates on how to make an agent master several RL tasks only with a single learned model. Linking pre-knowledge towards any tasks, the agent is greatly confused by the diversity of state representations and strategies. In short, multi-task generalization in RL mainly aims at learning a generalized model across diverse tasks. In this direction, previous works have been proposed with novel models and algorithms that are more capable of generalizing and transferring with shared representations. For example, the latest literature \cite{borsa2016learning} verifies the practicability of learning shared representations of value functions. However, it has been demonstrated that a completely shared model performs poorly \cite{romoff2016deep}. Furthermore, to deal with the problem of \textit{task diversity}, the agent in \cite{romoff2016deep} is designed to have different hidden layers and output layers for each task, remaining shared convolutional layers across all tasks. However, it can only learn to master 3 Atari games. An additional network \cite{sermanet2016unsupervised} has been designed to detect which task is the agent facing, and then to favor or select a corresponding stream in the model.

In summary, recent works have made great advances in MT-RL. However, they still use conventional deep neural network architecture, which does not consider \textit{hierarchical shared knowledge} across multiple tasks. This leads to poor generalization ability in mastering diverse tasks, which can be solved by the proposed GTN architecture presented in the following.

\section{Generalization Tower Network}

In this section, we introduce our GTN approach for MT-RL, which learns to master multiple RL tasks in a single learned model.
Specifically, the architecture of GTN is proposed at first.
Then, we present how GTN is implemented to work on multi-task scenarios in an asynchronous way, so that all tasks go through the same GTN model without requiring any task labels.

\subsection{Architecture of GTN}
\label{section-archi-of-gtn}

Figure \ref{figure-main-framework} shows the architecture of GTN.
We can see from this figure that the input to GTN is frame content from the game emulator, and the output is state value $V$ and policy $\pi$.
Here, $\pi$ is the set of vectors with different size, corresponding to the action space of different tasks.
As shown in Figure \ref{figure-main-framework}, GTN is a two-dimensional network, including horizontal and vertical streams.
In the horizontal stream, representations features are extracted ``layer by layer'', the same as the traditional deep neural networks. To model \textit{hierarchical shared knowledge}, the vertical stream is introduced in GTN to master multiple tasks, in which higher levels\footnote{In this paper, we use ``level'' and ``layer''to denote information flows of the vertical and horizontal streams in GTN, respectively. } learn more representations features than lower ones.
As such, our GTN architecture is suitable for diverse tasks, which can be well mastered by different numbers of features.
In the following, we present the horizontal and vertical streams of our GTN architecture, respectively.

\paragraph{Horizontal streams with multi-layers.}
To present the horizontal streams of GTN, we take the horizontal stream at the \textit{level 1} in (Figure \ref{figure-main-framework}) as an example.
The horizontal stream of GTN contains multiple convolutional layers, i.e., $C_{(1,1)}\rightarrow C_{(1,2)}\rightarrow,...,\rightarrow C_{(1,n)} \rightarrow,...,\rightarrow C_{(1,N)}$, to learn features from the high-dimensional input. Thus, the horizontal stream contains layers $\{1, \cdots, n, \cdots, N\}$.
Finally, the convolutional layer is flattened into a vector with flatten layer ($F$), followed by a single LSTM layer ($L_1$) to model the temporal correlation of RL tasks.
For the LSTM chunk, we use the standard LSTM \cite{hochreiter1997long} which includes input, forget and output gates.
Thus, the horizontal stream of our GTN mainly produces traditional representations features of deep convolutional neural network \cite{simonyan2014very} and learns temporal information with LSTM, the same as \cite{hausknecht2015deep}. Similar structure holds for horizontal streams at other levels.
\begin{figure*}[htb]
	\begin{center}
		\centerline{\includegraphics[width=1\linewidth]{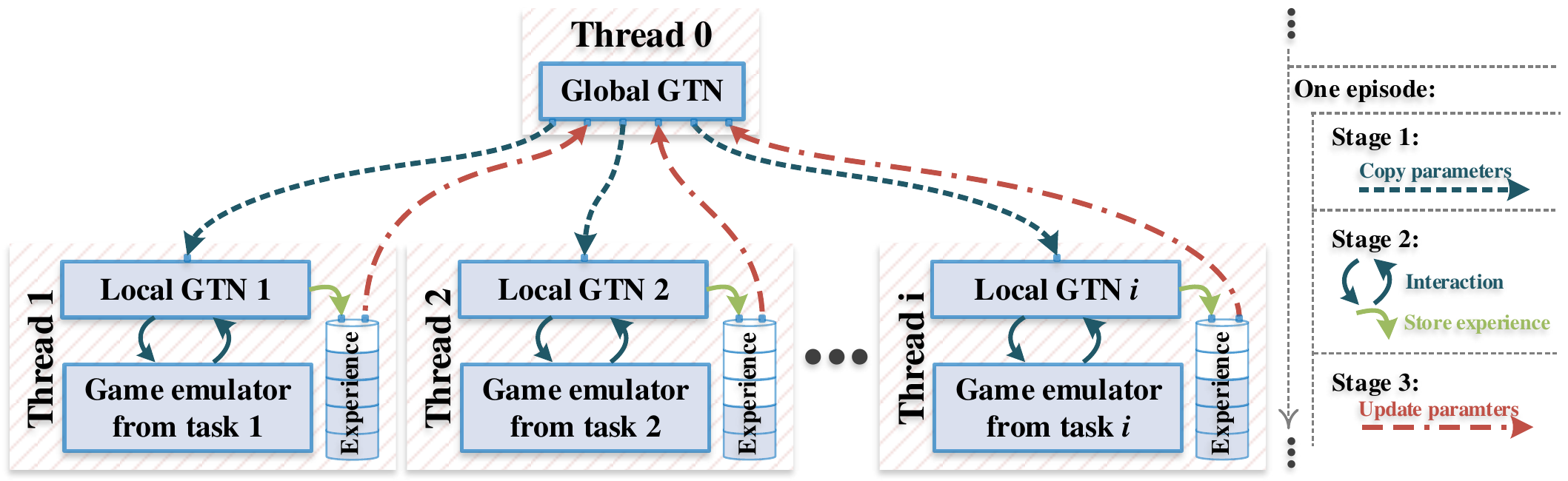}}
		\caption{Mastering multiple tasks with a single GTN model: implement GTN asynchronously.}
		\label{figure-asyn-mtrl-framework}
	\end{center}
\end{figure*}
\begin{figure*}[htb]
  \centering
    \subfigure[]{
    \label{figure-generalize-analysis-task}
    \includegraphics[width=0.45\textwidth]{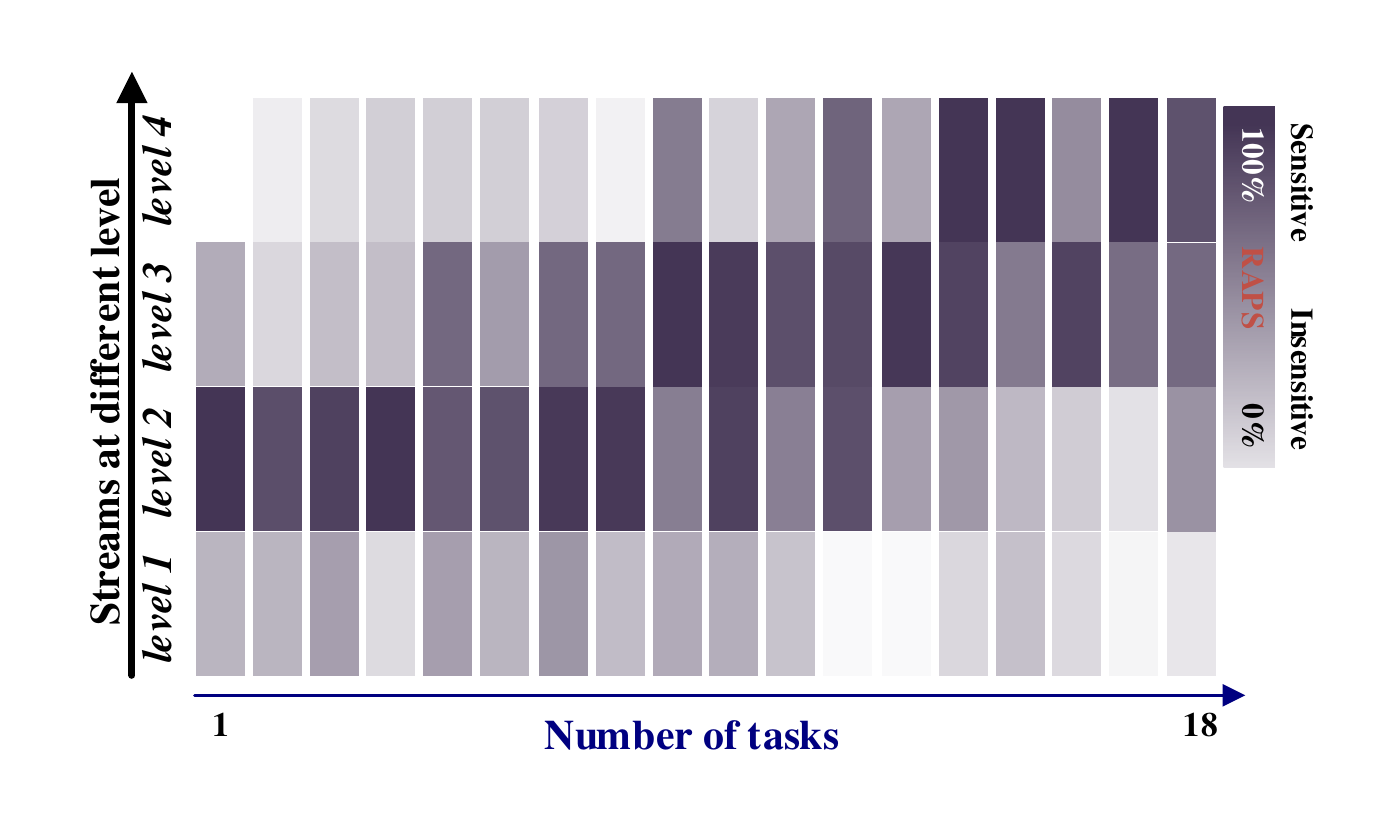}}
  \subfigure[]{
    \label{figure-generalize-analysis-epi}
    \includegraphics[width=0.45\textwidth]{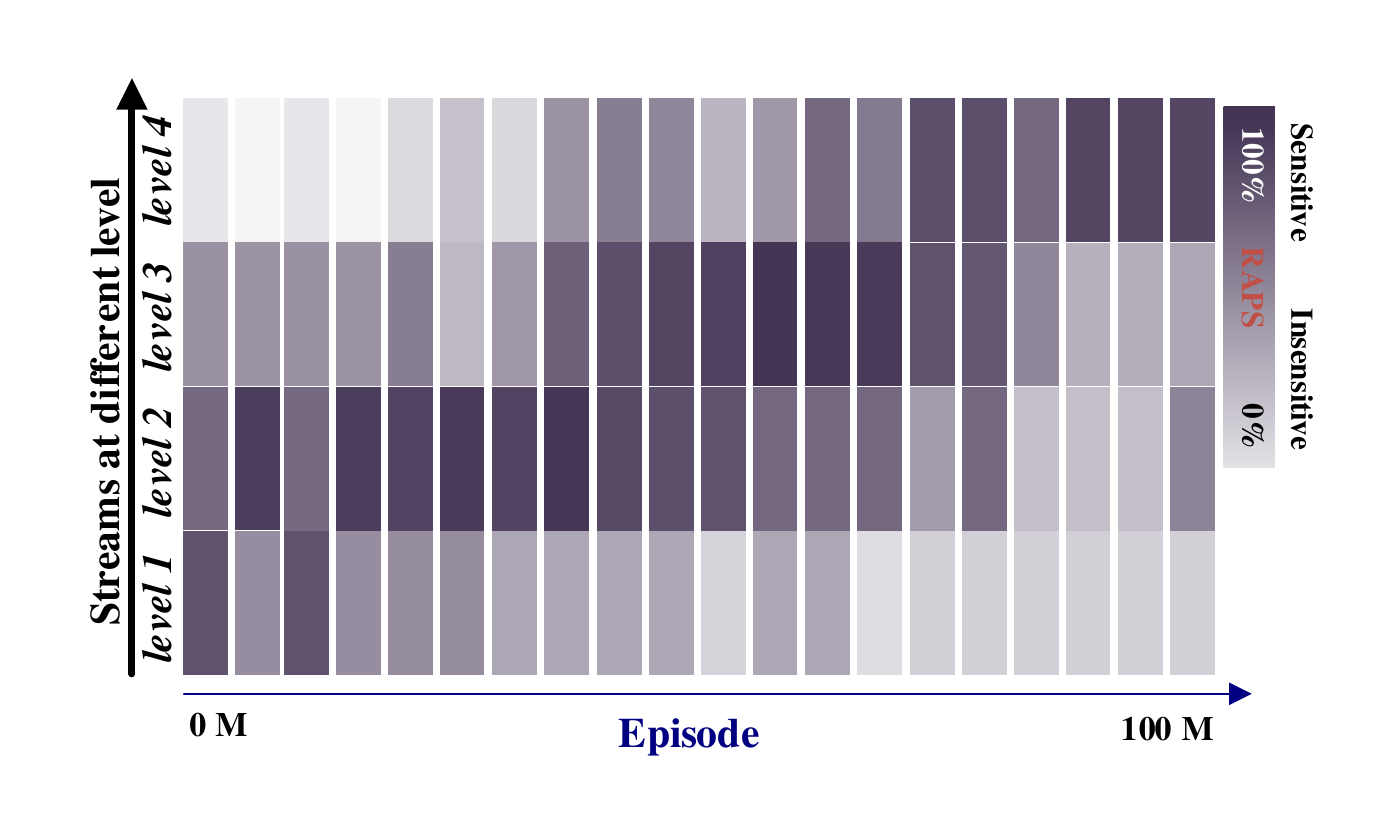}}
  \caption{
    (a): RAPS variance at different number of tasks.
    For obtaining RAPS, the experiment was conducted on a series of GTNs with $N=4$ and $M=4$, learning to master 1 to 18 \textit{shooting games} from Atari. The results were obtained at 100M training episodes.
    (b): RAPS variance alongside training episodes, obtained in learning all 18 \textit{shooting games}.
  }
\end{figure*}
\paragraph{Vertical stream with multi-levels.}
Our GTN architecture includes the vertical streams with levels $\{1, \cdots, m, \cdots, M\}$.
Specifically, after the first convolutional layer at \textit{level 1}, i.e., $C_{(1,1)}$, the stream of GTN splits into two sub-streams.
One continues to go through the remaining horizontal layers in \textit{level 1}, while the other is connected to \textit{level 2}, acting as the input to \textit{level 2}.
As such, every \textit{level  m} uses the feature map from the first convolutional layer at \textit{level m-1} as input.
This way, the vertical stream is formed in GTN.
Then, the last LSTM layers of the multiple horizontal streams, i.e.,  $L_1$, $L_2$, $\cdots$, $L_m$, $\cdots$, $L_M$ are combined together for yielding the final output.
Let the activation of LSTM layer $L_m$ be $a_m$.
The activation of the concatenation layer ($H$) is obtained by the Rectified Linear Units (\textrm{ReLU}) operation $\textrm{ReLU}=\max(0,x)$, as follows,
\begin{equation}
\label{combine}
H=\textrm{ReLU}(\sum^{M}_{m=1}a_{m}\textbf{T}_{m}),
\end{equation}
where $\{ \textbf{T}_1, \cdots \textbf{T}_m \cdots, \textbf{T}_M\}$ are learned parameters, to concatenate $M$ horizontal streams into a single layer.
Note that $\textbf{T}_{m}\in\mathbb{R}^{S \times A}$, where $S$ is the size of LSTM layer and $A$ is the size of concatenation layer.
The vertical stream aims to model \textit{hierarchical shared knowledge}, making our GTN fit to different tasks.

\subsection{Implementation of GTN in asynchronous method}
\label{section-asyn-mtrl-framework}

To learn multiple tasks with a single GTN model, we develop an asynchronous method to implement GTN based on \cite{mnih2016asynchronous}.
Our asynchronous method is summarized in Figure \ref{figure-asyn-mtrl-framework}.
Specifically, there is one thread (thread $0$) holding a global GTN and each of other multiple threads (thread $1,..., i$) holding a local GTN and a game emulator.
Note that both global and local GTNs use the same architecture as depicted in Figure \ref{figure-main-framework}.
In a single episode for each thread, the following procedure is conducted.
\begin{itemize}
    \item  Stage 1: The thread copies all parameters from global GTN to its own local GTN.
    \item  Stage 2:  The local GTN interacts with the game emulator in the same thread and stores the experiences.
    \item  Stage 3:  Once gathering enough experiences (i.e., meet terminal condition $t_{max}$), the thread updates all parameters in the global GTN via accumulating gradient.
\end{itemize}

At Stage 2, the local GTN and the game emulator interact with each other for several steps $\{1,\cdots t \cdots, t_{max}\}$.
At step $t$, GTN obtains \textit{observation} $o_t$ from the game emulator, which is the frame content gray-scaled and down-sampled to $42 \times 42$.
Then, observation $o_t$ is provided to GTN, together with the LSTM feature from the GTN at last step $f_{t-1}$.
Afterwards, GTN produces a set of policies with different size, corresponding to the size-varied actions of MT-RL.
Given policy $\pi$, action $a_t$ can be generated with standard deviation $\varepsilon$ to ensure exploration.
According to $a_t$, the game emulator updates from $o_t$ to $o_{t+1}$, returning a step reward $r_t$.
At the end of step $t$, a set of experiences $\{o_t,f_{t-1},a_t,r_t\}$ is stored for updating parameters at Stage 3.
At Stage 3, we apply the accumulating gradient to update parameters as follows. We assume that $\theta$ and $\theta_v$ are parameter vectors that generate policy $\pi$ and state value $V$, in global GTN. In local GTN, $\theta'$ and $\theta'_v$ are parameter vectors for $\pi$ and $V$. Then, the accumulating gradients are calculated as
\begin{equation}
\label{acc grad 1}
d\theta \leftarrow d \theta + \sum_{t=1}^{t_{max}} \nabla_{\theta'}\log\pi(a_t|o_t,f_{t-1};\theta')(R_t-V(o_t,f_{t-1};\theta'_v)),
\end{equation}
\begin{equation}
\label{acc grad 2}
d\theta_v \leftarrow d \theta_v + \sum_{t=1}^{t_{max}} \partial(R_t-V(o_t,f_{t-1};\theta'_v))^2/\partial\theta'_v.
\end{equation}
In the above equation, $R_t$ denotes the discounted reward:
 \begin{equation}
\label{R}
R_{t}=\sum_{i=t}^{t_{max}} \gamma^{i-t} r_{i}.
\end{equation}
where $r_i$ is the step reward, and $\gamma$ (=0.99) is the discount factor for future rewards.

\section{Generalization Analysis}
\label{section-generalization-analysis}

Now, we analyze the generalization ability of our GTN approach across multiple tasks. In this paper, generalization ability refers to the performance of GTN, when mastering different numbers of tasks within a single learned model.
Since our GTN introduces the vertical stream for learning multiple tasks, we investigate the generalization ability of GTN at different levels of the vertical stream.
For such investigation, we measure the average perturbation sensitivity (APS) \cite{rusu2016progressive}, which indicates how much the final output relies on a specific level of GTN.
To obtain APS, we first inject Gaussian noise $\mathcal{N}(0,1)$ to a specific horizontal stream at one level for producing perturbation.
Then, we measure the impact of this perturbation on the performance, using APS.
In this paper, APS is measured by the percentage of score dropping, due to the noise perturbation. We compute the relative APS (RAPS) via dividing APS by the sum of the APSs of all streams, to investigate the relative dependency of different levels of GTN.

\begin{figure*}[htb]
	\begin{center}
		\centerline{\includegraphics[width=0.92\textwidth]{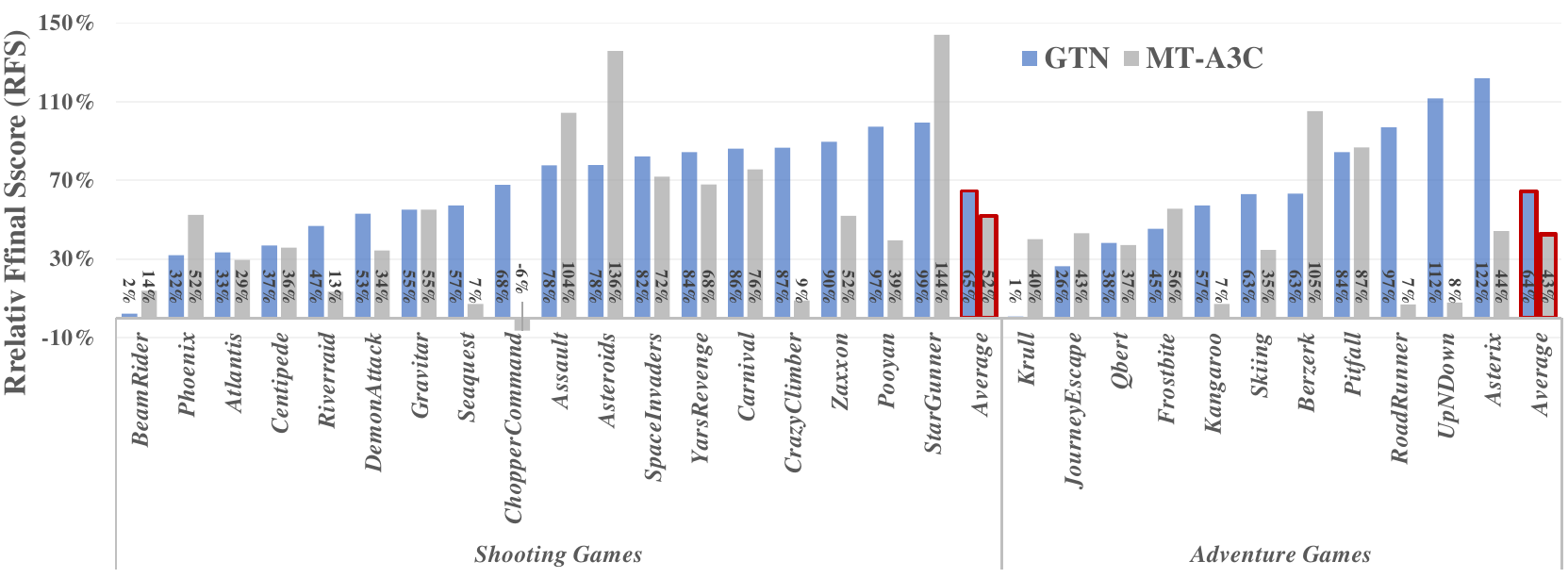}}
		\caption{
                Comparison of RFS over classified Atari games: \textit{Shooting Games} and \textit{Adventure Games}.
            }
		\label{figure-resn-1}
	\end{center}
\end{figure*}

Figure \ref{figure-generalize-analysis-task} shows the RAPS results of different levels in GTN, when learning 1 to 18 tasks.
The results are obtained by running our GTN approach, averaged over all \textit{shooting games} of Atari as list in Figure \ref{figure-resn-1}.
From Figure \ref{figure-generalize-analysis-task}, we can see that RAPS of high-level (e.g., levels 3 and 4) is increased at more tasks.
This indicates that when learning to master more tasks, GTN turns to rely on higher levels.
Thus, the proposed architecture of GTN, especially the vertical stream, is effective in improving the generalization ability for MT-RL.
Figure \ref{figure-generalize-analysis-epi} further draws the RAPS results of different levels versus training episodes, when running our GTN approach to handle all \textit{shooting games} of Atari.
One may observe that RAPS can be increased at more training episodes, for high-level in GTN (e.g., levels 3 and 4).
By contrast, RAPS of low-level decreases alongside increased episodes.
In a word, more levels in the vertical stream and larger number of training episodes are required for improving the generalization ability for MT-RL.

\begin{figure*}[htb]
	\begin{center}
		\centerline{\includegraphics[width=0.92\textwidth]{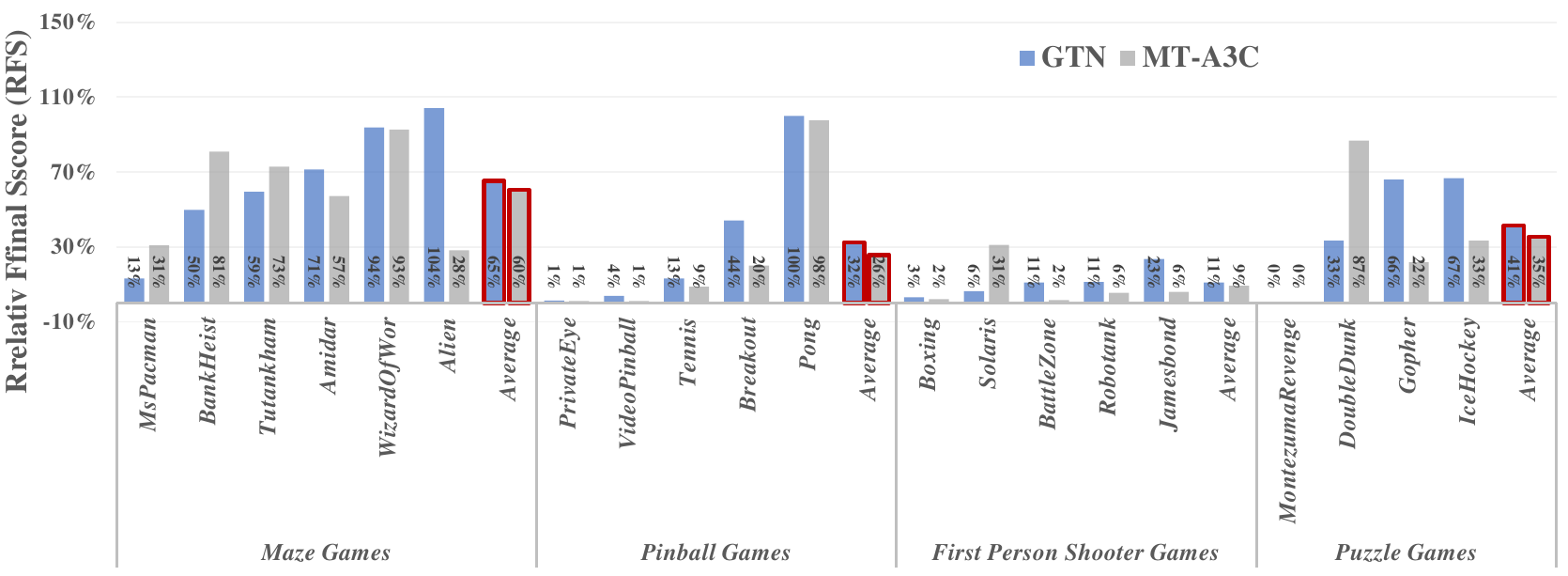}}
		\caption{
                Comparison of RFS over classified Atari games: \textit{Maze Games}, \textit{Pinball Games}, \textit{First Person Shooter Games} and \textit{Puzzle Games}.
            }
		\label{figure-resn-2}
	\end{center}
\end{figure*}

\begin{figure*}[htb]
  \centering
  \subfigure[]{
    \label{figure-ablative-study-s}
    \includegraphics[width=0.43\textwidth]{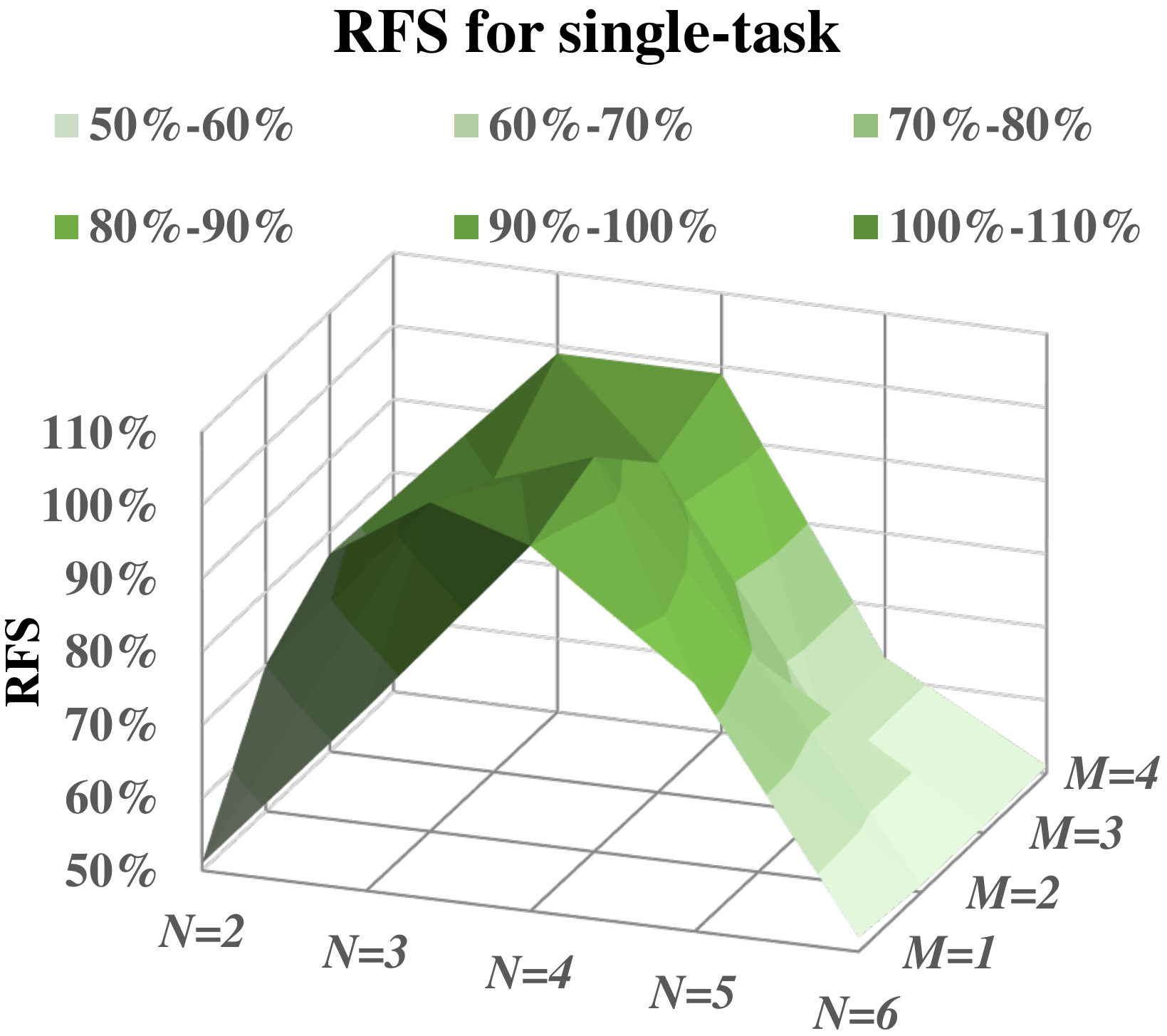}}
  \subfigure[]{
    \label{figure-ablative-study-m}
    \includegraphics[width=0.43\textwidth]{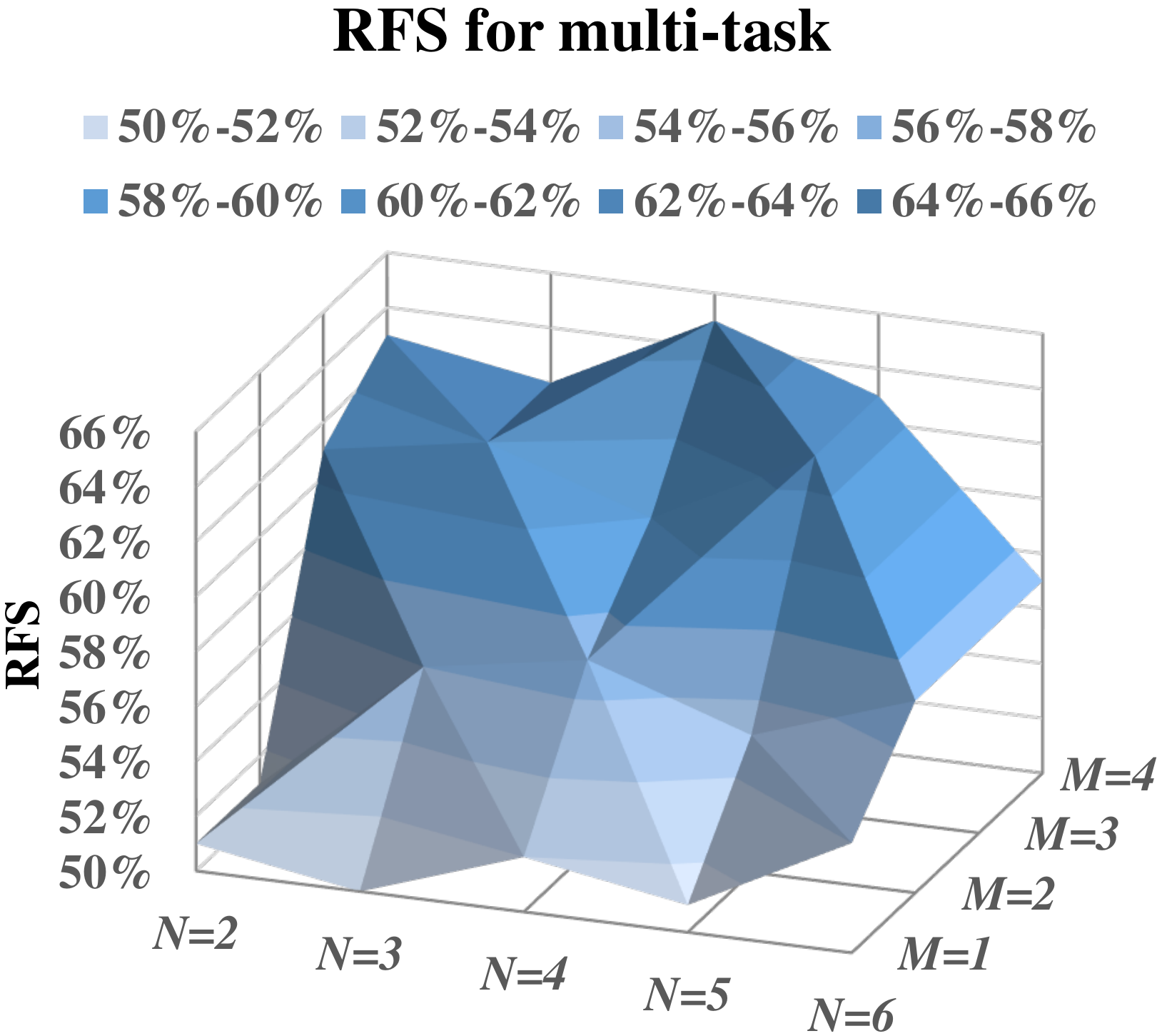}}
  \caption{RFS results of our GTN approach at different $M$ and $N$ for mastering (a) single task and (b) multiple tasks. Note that the RFS results are averaged over all \textit{shooting games} of Atari. For single task, each game is individually played by \textit{agent} of GTN. For multiple tasks, all games are synchronously played by \textit{agent} of GTN.
  }
\end{figure*}

\section{Experiment}

\subsection{Setup}
\label{section-exp-setup}

In our experiments, we manually divide the Atari games into 6 categories, according to the similarity of tasks. All tasks belonging to each category can be found from Figures \ref{figure-resn-1} and \ref{figure-resn-2}. They cover 51 of total 57 Atari games, and the remaining 6 games are hard to be classified. Thus, our experiments were conducted on those 51 games.

\paragraph{GTN Setup.}
The GTN is simply set with 4 levels ($M=4$) in vertical streams and 4 layers ($N=4$) in horizontal streams.
In our GTN, the architecture of each horizontal stream is the same as that of A3C \cite{mnih2016asynchronous}, for fair comparison.
Specifically, for all convolutional layers, the kernel size is $3 \times 3$ and the number of kernels is 32.
The stride during convolution is set to 2.
Besides, the size of the LSTM layers is all set to 288.
For playing the Atari games, the minimal and maximal numbers of \textit{actions} are 5 and 18, respectively.
To master diverse tasks in a unified form, during the first 2 episodes, the agent detects and records the maximal step reward.
Afterwards, every step reward is normalized by dividing this stored maximal step reward.
We follow \cite{mnih2016asynchronous} to configure other unmentioned settings.

\paragraph{Evaluation Metric.} To evaluate the multi-task generalization ability, we measure how much a multi-task agent achieves or exceeds the scores performed by the corresponding single-task agent. To this end, we measure the relative final scores (RFS):
\begin{equation}
\label{eva}
\textrm{RFS}=\frac{\text{Score}_{\text{multi-task}}}{\text{Score}_{\text{single-task}}},
\end{equation}
which eliminates the diverse scales of scores across multiple tasks\footnote{Since the baseline of raw scores from different tasks is diverse, we follow \cite{mnih2015human} to subtract the corresponding scores of random actions from the raw scores.}. Note that the same metric is also used in most recent works of MT-RL \cite{rusu2015policy,parisotto16_actormimic}.

\subsection{Performance Analysis of GTN}
\label{section-ablative-study}
Since the vertical and horizontal streams are the main contribution of the proposed GTN architecture, we analyze the performance of our GTN approach, with different numbers of levels ($M$) and layers ($N$).
In the following, the performance of \textit{agent} is evaluated on single task and multiple tasks, when $M$ ranges from 1 to 4 and $N$ varies from 1 to 6.

Figure \ref{figure-ablative-study-s} shows the performance of GTN on single task, along with increased $N$ and $M$.
The performance in this figure is measured by RFS, averaged over all \textit{shooting games} of Atari.
We can see from this figure that when making GTN work on single task, the averaged RFS dramatically changes at different $N$ and it can achieve $104\%$ at $N=4$ and $M=2$.
In contrast, $M$ slightly improves the performance of our GTN on single task.
This implies the horizontal stream is more effective in handling single task in RL.
For multiple tasks, the averaged RFS values of playing all \textit{shooting games} are used for performance evaluation.
Figure \ref{figure-ablative-study-m} shows the averaged RFS values of GTN on handling multiple tasks of all \textit{shooting games}, with different $N$ and $M$.
We can find from this figure that enhancement of $M$ from 1 to 4 improves the averaged RFS from $52\%$ to $65\%$ at $N=4$, whereas $N$ has little impact on the averaged RFS.
In a word, the above analysis indicates that the horizonal stream in our GTN approach is effective in mastering single task, and the vertical stream is "good at'' learning multiple tasks.

\subsection{Evaluation on Performance.}

This section compares the RFS results between the proposed GTN and conventional MT-A3C \cite{romoff2016deep}, both based on \cite{mnih2016asynchronous}. In particular, MT-A3C means that the asynchronous MT-RL framework is implemented with the same traditional deep neural network architecture as A3C, instead of our GTN.
It is worth mentioning that we do not compare against progressive network \cite{rusu2016progressive}. It is because progressive network \cite{rusu2016progressive} requires manual inference to select corresponding stored model for different tasks, whereas our GTN approach requires no task labels and enables mastering multi-tasks in one model. In other words, it is unfair to compare our approach with \cite{rusu2016progressive}.

Figures \ref{figure-resn-1} and \ref{figure-resn-2} plot the RFS results of MT-A3C and GTN for each game of different categories.
We can see from these figures that GTN performs better than MT-A3C in all 6 categories, with up to \textbf{23\%} improvement (\textit{shooting games}) in average RFS.
Interestingly, the performance gap between GTN and MT-A3C is especially large for \textit{Shooting Games} and \textit{Adventure Games}, which have more tasks than other 4 categories.
This indicates that our GTN approach is more suitable for MT-RL with larger number of similar RL tasks, thus showing the high generalization ability of our GTN approach.
This also makes our approach practical in real-world application.

\section{Conclusion}

This paper has proposed a novel deep neural network architecture called GTN for MT-RL, which incorporates both horizontal and vertical streams within a single learned model.
Different from the conventional MT-RL approaches, the vertical stream proposed in our GTN architecture enables better performance when dealing with multiple tasks.
This is mainly motivated by the fact that multiple tasks may share common knowledge in hierarchy.
Experimental results also showed that our GTN approach improves the performance of A3C in MT-RL, over 51 Atari games.

\clearpage
\bibliographystyle{aaai}
\bibliography{aaai2018_gtn}

\end{document}